\documentclass[letterpaper, 10 pt, conference]{ieeeconf}
\IEEEoverridecommandlockouts
\usepackage{amssymb}
\usepackage{gensymb}
\usepackage{stmaryrd}
\usepackage{cite}
\usepackage{color}
\usepackage{multirow}
\usepackage{graphicx,times}
\usepackage{epstopdf}
\usepackage{indentfirst}
\usepackage{CJK}
\usepackage{amsmath}
\usepackage{amsfonts}
\usepackage{txfonts}
\usepackage{mathrsfs}
\usepackage{theorem}
\usepackage{float}
\usepackage{subfigure}
\usepackage{booktabs}
\usepackage{verbatim}
\usepackage{hhline}     		
\usepackage{algpseudocode}
\usepackage{algorithm}
\usepackage{array}
\usepackage{url}
\usepackage{bm}
\usepackage{xcolor}
\usepackage{microtype}
\usepackage{cleveref}
\usepackage{balance}

\newcommand{\ie}{{\em i.e.}}
\newcommand{\eg}{{\em e.g.}}
\newcommand{\et}{{\em et al.}}

\def\spth{\textsuperscript{th}}

\def\BibTeX{{\rm B\kern-.05em{\sc i\kern-.025em b}\kern-.08em
    T\kern-.1667em\lower.7ex\hbox{E}\kern-.125emX}}
\begin{document}

\title{VISC: mmWave Radar Scene Flow Estimation using Pervasive Visual-Inertial Supervision
\author{Kezhong Liu$^{1}$, Yiwen Zhou$^{1}$, Mozi Chen$^{1}$, Jianhua He$^{2}$, Jingao Xu$^{3}$, Zheng Yang$^{4}$, \\Chris Xiaoxuan Lu$^{5}$, Shengkai Zhang$^{1}$$^{*}$
}
\thanks{Authors$^{1}$ are with State Key Laboratory of Maritime Technology and Safety, Wuhan University of Technology, Wuhan, China. {\tt\small \{kzliu, zyw293423, shengkai, chenmz\}@whut.edu.cn}}

\thanks{Author$^{2}$ is with University of Essex, Colchester, UK. {\tt\small j.he@essex.ac.uk}}

\thanks{Author$^{3}$ is with Carnegie Mellon University, Pittsburgh, USA. {\tt\small jingaox@andrew.cmu.edu}}

\thanks{Author$^{4}$ is with Tsinghua University, Beijing, China. {\tt\small hmilyyz@gmail.com}}

\thanks{Author$^{5}$ is with University College London, London, UK. {\tt\small xiaoxuan.lu@ucl.ac.uk}}

\thanks{$^{*}$Corresponding author: Shengkai Zhang (shengkai@whut.edu.cn)}

}


\maketitle

\begin{abstract}
This work proposes a mmWave radar's scene flow estimation framework supervised by data from a widespread visual-inertial (VI) sensor suite, allowing crowdsourced training data from smart vehicles. Current scene flow estimation methods for mmWave radar are typically supervised by dense point clouds from 3D LiDARs, which are expensive and not widely available in smart vehicles. While VI data are more accessible, visual images alone cannot capture the 3D motions of moving objects, making it difficult to supervise their scene flow. Moreover, the temporal drift of VI rigid transformation also degenerates the scene flow estimation of static points. To address these challenges, we propose a drift-free rigid transformation estimator that fuses kinematic model-based ego-motions with neural network-learned results. It provides strong supervision signals to radar-based rigid transformation and infers the scene flow of static points. Then, we develop an optical-mmWave supervision extraction module that extracts the supervision signals of radar rigid transformation and scene flow. It strengthens the supervision by learning the scene flow of dynamic points with the joint constraints of optical and mmWave radar measurements. Extensive experiments demonstrate that, in smoke-filled environments, our method even outperforms state-of-the-art (SOTA) approaches using costly LiDARs. 

\end{abstract}
\label{sec:intro}    
\section{Introduction}

Millimeter-wave (mmWave) radar, owing to its radio-fre-quency (RF) nature, has been indispensable in environmental perception against adverse weather and lighting conditions for smart vehicles, \eg, drones and cars. However, its limited sensing resolution, \ie, sparse and noisy point clouds, poses challenges in scene flow estimation for safe navigation. Currently, the mainstream is to learn the radar scene flow using the supervision of dense point clouds from 3D LiDAR~\cite{dong2022exploiting, baur2021slim, chen2022gocomfort} since its direct 3D observations can easily track the 3D motion state of moving objects. Despite their high accuracy, such a learning-based paradigm requires a large scale of training data to improve model generalization. 

Since high-performance 3D LiDAR for automobile is costly and thus not commonly available in smart vehicles, learning the radar scene flow via visual data from low-cost and pervasively available cameras would be more favorable. In fact, current visual-inertial SLAM implementations~\cite{qin2018vins, ORBSLAM3_2021} can translate the vehicle's rigid transformation to the scene flow of static points. However, it is difficult to track the scene flow of dynamic points due to the lack of 3D observations~\cite{zhou2022integrated}. 



\begin{figure}[h]
	\centering
	\includegraphics[width=3.2in]{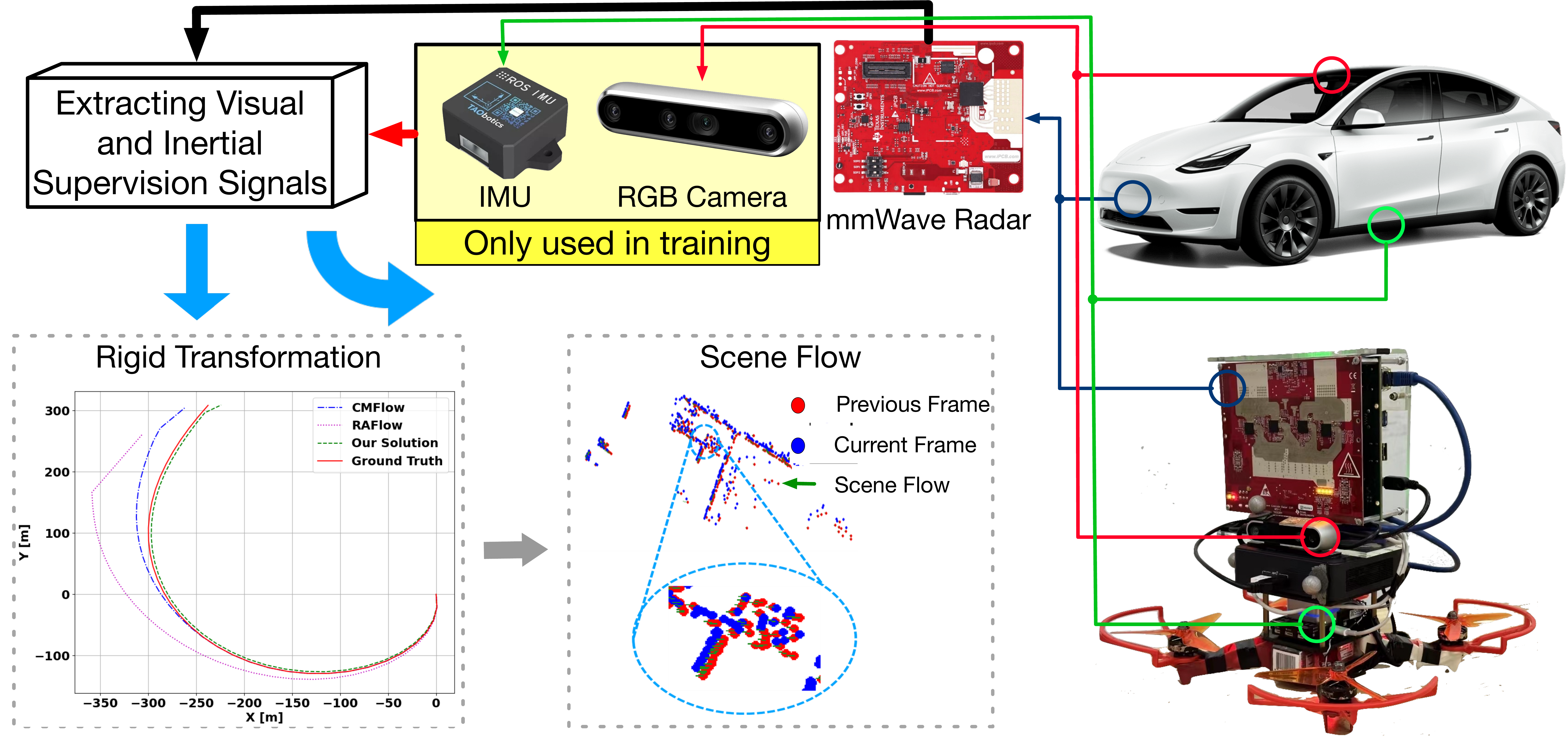}
	\caption{Our system extracts the supervision signals from visual and inertial data to learn the mmWave radar scene flow model. The images and inertial measurements are only required in the training stage.}
	\label{fig:toy}
	\vspace{-3mm}
\end{figure}

This paper presents VISC, a mmWave radar scene flow estimation framework for smart vehicles guided by a pervasive visual-inertial (VI) sensor suite, as shown in Fig.~\ref{fig:toy}, supervising the motion state of moving objects in the lack of 3D/depth observations. As annotating point clouds from mmWave radar is challenging, our design follows a self-supervised learning paradigm~\cite{ding2022self, ding2023hidden}. VISC only takes images and inertial measurements during the training. {\em In the test, it estimates the scene flow using only the radar's point clouds, working in smoke-filling environments.} The opportunity to supervise the motion of moving objects with visual images is to exploit the tight coupling of scene flow and rigid transformation. For example, the VI sensor suite can estimate accurate rigid transformation through multi-view constraints~\cite{qin2018vins, ORBSLAM3_2021}. The rigid transformation constrains the scene flow of static points. For dynamic points, the radar's relative radial velocity (RRV) and the camera's optical flow contribute partial observations to their scene flow. 

While the above examples are intuitive, extracting the supervision signals encounters two technical challenges. One is that VI rigid transformation suffers from temporal drift without loop closure or GNSS, failing to provide trustworthy supervisions to guide the radar sensing. In addition, the rigid transformation error will translate into the scene flow estimation of static points. Our design aims to enable radar scene flow in both indoor and outdoor environments for various smart vehicles, \eg, cars and drones. Therefore, assuming the availability of loop closure and GNSS readings is unreasonable. The other challenge is to build the supervision signal extraction module for guiding the scene flow with dynamic points, whose 3D positions are partially observable.




VISC consists of two designs to address the above challenges. {\em First}, we propose a kinematic-learned sensor fusion method that compensates for the temporal drift of VI rigid transformation. Inspired by dead reckoning (DR)~\cite{herath2020ronin, liu2020tlio} for human motion patterns, we develop a deep neural network that learns a vehicle's 3D motion pattern valid for a segment of the IMU data. Through a random perturbation, the learned motion model becomes an independent odometer, and thus, fusing it with the kinematic VI rigid transformation can mitigate the temporal drift. {\em Second}, we develop an optical-mmWave supervision extraction method that extracts the supervision signals of radar rigid transformation and scene flow. It strengthens the supervision by learning the scene flow of dynamic points with joint constraints of optical flow and VI dynamic 3D reconstruction~\cite{fan2024enhancing}. 




{\bf Contributions.} We enable the mmWave radar scene flow estimation using a pervasive VI sensor suit by 1) a kinematic-learned sensor fusion method that fuses a vehicle's network-learned motion model with kinematic VI model to compensate for the temporal drift of VI rigid transformation, and 2) an optical-mmWave supervision extraction method that extracts the supervision signals of radar rigid transformation and scene flow. We collect data from the Carla simulator and our customized sensor platform for extensive experiments. The results show that VISC can work in smoky environments and outperforms the SOTA solutions using high-cost LiDARs.


\label{sec:related}
\section{Related Work}

{\bfseries Scene flow by camera.} Since monocular scene flow is a highly ill-posed problem~\cite{hur2020self}, vision-based approaches either use depth images from RGB-D~\cite{lv2018learning} and stereo cameras~\cite{jiang2019sense}, or monocular images with given depth estimates~\cite{hur2020self}. Large image datasets have made supervised approaches~\cite{jiang2019sense} highly accurate. But they may suffer from domain overfitting. To avoid domain overfitting and save the trouble of obtaining ground-truth data, unsupervised approaches~\cite{lee2019learning} have been proposed, however, with inferior performance. In addition, vision-based approaches are limited to good illumination conditions. Thus, LiDAR, which emits modulated light, has been another mainstream sensor for scene flow estimation.  

{\bfseries Scene flow by LiDAR.} 3D point clouds from LiDAR can infer point-wise scene flow directly. Currently, supervised learning approaches~\cite{wang2021festa, wei2021pv} have achieved SOTA accuracy. However, they require considerable effort to annotate the scene flow data manually. Thus, their self-supervised counterparts~\cite{li2022rigidflow} are developed to train models on unannotated data. As expected, the performance of self-supervised methods is limited due to the absence of real labels. Albeit the promising results of these methods, LiDAR is still incapable of dealing with adverse weather, \eg, fog, storms, and downpours, for smart vehicles.

{\bfseries Scene flow  by mmWave radar.} The millimeter-level wavelength makes mmWave radar suffer from a low environmental resolution, manifesting sparse point clouds~\cite{zhang2024waffle, liu2024pmtrack}. Meanwhile, due to the difficulty of annotating sparse point clouds, existing solutions prefer self-supervised learning approaches~\cite{ding2023hidden, ding2022self}. Ding~\et~\cite{ding2023hidden} proposed CMFlow, representing the SOTA radar's scene flow estimation approach. Its superior performance comes from the cross-modal supervision retrieved by redundant colocated sensors, including camera, high-performance LiDAR, and RTK-GPS. Among them, a high-performance LiDAR for autonomous driving costs several thousand US dollars, and the RTK-GPS relies on a reference station network, which is not widely available. In contrast, our proposal aims to learn the radar scene flow using a low-cost VI sensor suite, which is commonly installed on commercial vehicles.


\label{sec:design}    
\section{Design of VISC}
\label{sec:design}

\begin{figure}[h]
	\centering
	\includegraphics[width=3.3in]{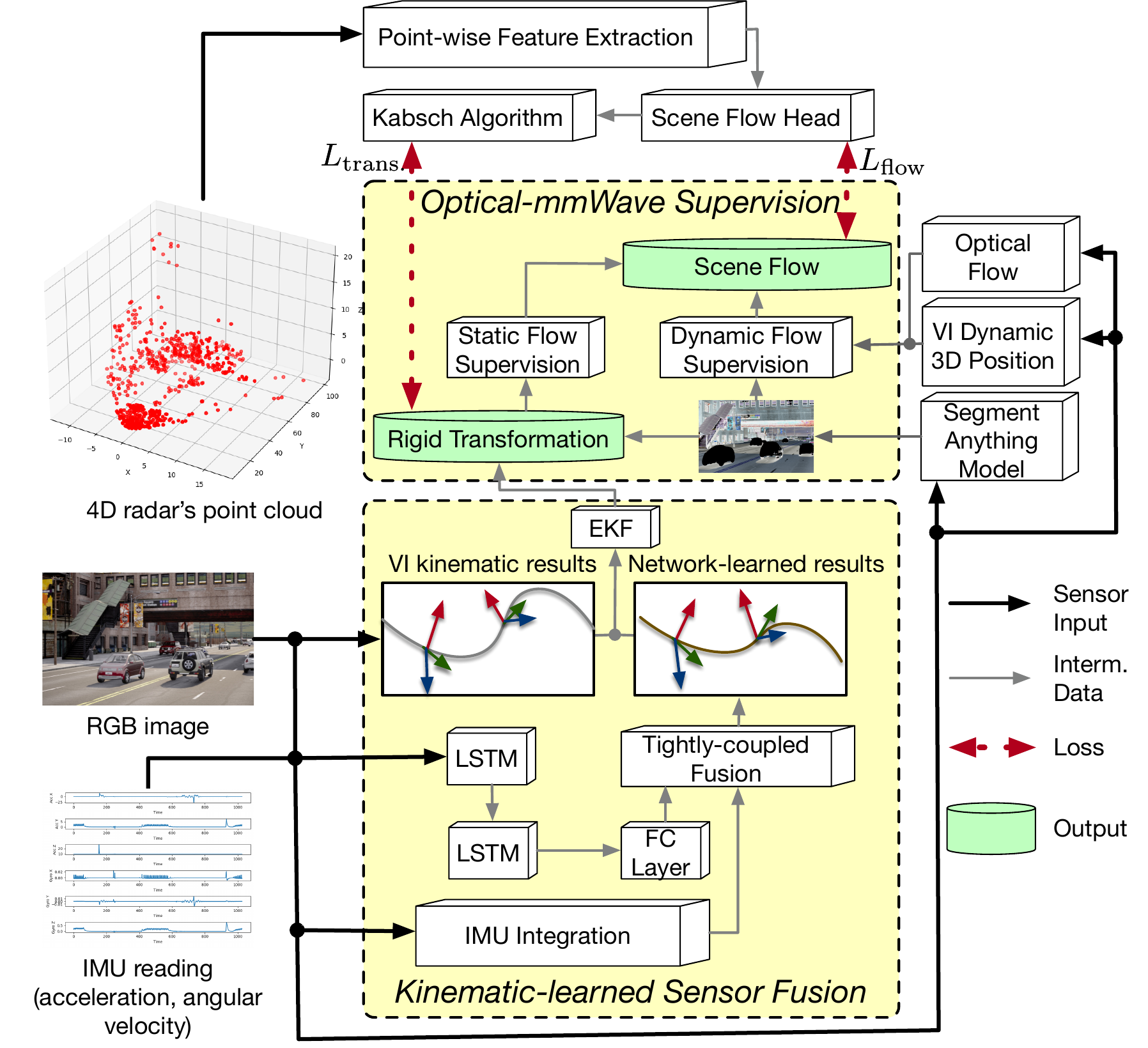}
	\caption{The system overview of VISC. Our new designs are highlighted in the yellow boxes.}
	\label{fig:overview}
	\vspace{-3mm}
\end{figure}

\subsection{Overview}
\label{subsec:overview}


Fig.~\ref{fig:overview} shows the system overview, which consists of a {\em kinematic-learned sensor fusion} module and a {\em optical-mmWave supervision extraction} module. The inputs include RGB images, IMU readings, and point clouds from a mmWave radar. We first extract point-wise latent features from two consecutive point clouds~\cite{ding2022self}. Then, they are sent to two heads to infer the radar rigid transformation and coarse scene flow~\cite{ding2023hidden}. 

To supervise the above coarse estimates, we first estimate centimeter-level rigid transformation from the kinematic-learned sensor fusion module (refer to~\S\ref{subsec:fusion}) to provide drift-free rigid transformations. It holds the supervisions of radar-based rigid transformation and the scene flow of static points, which can be masked out by existing large foundation models for object segmentation, \eg, Segment Anything Model~\cite{kirillov2023segany}. Then, the optical-mmWave supervision extraction module extracts the supervision signals of radar rigid transformation and scene flow. Finally, we jointly optimize the radar-based rigid transformation and scene flow. 

\subsection{Kinematic-learned Sensor Fusion}
\label{subsec:fusion}



The obtained rigid transformation from VI-SLAM~\cite{qin2018vins} has already come from the visual-inertial fusion. If we would like to mitigate the temporal drift further, a third sensor would be required. We can achieve this goal without adding another sensor. IMU-based DR solutions~\cite{herath2020ronin, liu2020tlio} have demonstrated the IMU's capability of being an independent sensor for rigid transformation estimation. 

\subsubsection{Inertial-learning Deep Neural Network}
\label{subsubsec:network}

{\bf Network architecture.} Since an IMU provides a sequence of 3D accelerations and angular velocities to infer the rigid transformation over time, we choose the special recurrent neural network (RNN) structure, namely Long Short-Term Memory (LSTM), for sequence modeling. We employ the full-connected LSTM (FC-LSTM) due to its power to handle temporal correlation~\cite{graves2013generating}. Specifically, the network consists of two LSTM layers and one fully-connected layer. We deploy two LSTM layers because deeper models can give better results while no significant performance improvement between the two-layer and three-layer networks~\cite{srivastava2015unsupervised}. 

Each LSTM layer contains $64$ units. We denote the hidden state of each unit as $\mathbf{L}_k$, $k = 1, 2, \dots, 64$. Each $\mathbf{L}_k$ takes IMU data $\hat{\bm{a}}_k$ and $\hat{\bm{\omega}}_k$, as well as the output of the previous unit for further processing. The output from the LSTM layers is mapped to a $1\times 12$ vector through the FC layer.

{\bfseries Loss function}. Since the network outputs 3D translations and uncertainties between two consecutive frames of images, we use the mean square error (MSE) loss and the Gaussian maximum likelihood (ML) loss during training.
\begin{equation}
	\mathcal{L}_\text{MSE}(\bm{t}, \hat{\bm{t}}) = \frac{1}{N}\sum_{i=1}^{N}\|\bm{t}_i - \hat{\bm{t}}_i\|^2,
\end{equation}
\begin{equation}
	\mathcal{L}_\text{ML}(\bm{t}, \hat{\bm{\Gamma}}, \hat{\bm{t}}) = \frac{1}{N}\sum_{i=1}^{N}\frac{1}{2}\left(\log\det(\hat{\bm{\Gamma}}_i)+\|\bm{t}_i - \hat{\bm{t}}_i\|^2\right),
\end{equation}
where $\bm{t} = \{\bm{t}_1, \bm{t}_2, \cdots, \bm{t}_N\}$ denotes the ground truth 3D translation provided by VI-SLAM. $N$ is the number of IMU segments divided by visual frames in the training dataset. $\hat{\bm{t}} = \{\hat{\bm{t}}_1, \hat{\bm{t}}_2, \cdots, \hat{\bm{t}}_N\}$ and $\hat{\bm{\Gamma}} = \{\hat{\bm{\Gamma}}_1, \hat{\bm{\Gamma}}_2, \cdots, \hat{\bm{\Gamma}}_N\}$ are the translations and corresponding uncertainties learned from the neural network. $\hat{\bm{\Gamma}}_i, 1 \leq i \leq N$, is a $3 \times 3$ covariance matrix for $t$-th data segment. Here we simply follow a diagonal covariance parameterization~\cite{liu2020tlio} by three coefficients $\hat{\bm{c}}_i = [\hat{c}_{ix}, \hat{c}_{iy}, \hat{c}_{iz}]^\top$ from the network output. Then $\hat{\bm{\Gamma}}_i$ can be written as $\hat{\bm{\Gamma}}_i = \text{diag}\left(e^{2\hat{c}_{ix}}, e^{2\hat{c}_{iy}}, e^{2\hat{c}_{iz}}\right)$.


\subsubsection{Two-stage Sensor Fusion}


{\bf Tightly-coupled fusion.} We adopt a graph-based optimization framework to fuse the outputs from the IMU kinematic and statistical model in a tightly-coupled fashion, which can typically achieve better accuracy~\cite{zhang2022lora, qin2018vins, zhang2020rf}.

Let $\mathbf{x}_i$ denote the vehicle state at time $i$. At each $i$, the deep neural network output $\hat{\mathbf{z}}_i$ which includes 3D translation $\hat{\bm{t}}_i$ and uncertainty $\hat{\bm{\Gamma}}_i$. Our system divides the IMU segment by image frames as the IMU rate ($100$ Hz) is higher than the camera frame rate ($20$ Hz). We integrate the measurements in the data segment to obtain relative state $\hat{\mathbf{u}}_{i+1}^i$ between two consecutive frames.


To achieve real-time processing, we maintain a sliding window of $m$ states in the optimization~\cite{zhang2020rf}. The full state vector in the optimization can be defined as $\bm{\mathcal{X}} = \left[ \mathbf{p}_1^0; \mathbf{p}_2^0; \cdots, \mathbf{p}_m^0\right]$, where $\mathbf{p}_i^0$ denotes $i$\spth position with respect to the first ($0$\spth) position. 
Our objective is to minimize the Mahalanobis norm of all measurement residuals to obtain a maximum a posteriori estimation: 
\begin{equation}
	\min_{\bm{\mathcal{X}}} \left\{\sum_{i\in\mathcal{N}}\left\| \hat{\mathbf{z}}_i - \mathbf{H}_i\bm{\mathcal{X}} \right\|_{\mathbf{P}_i}^2 + \sum_{j\in\mathcal{I}}\left\|\hat{\mathbf{u}}_{j+1}^j - \mathbf{H}_{j+1}^j \bm{\mathcal{X}} \right\|_{\mathbf{P}_{j+1}^j}^2\right\},
	\label{eqn:linear}
\end{equation}
where $\mathbf{H}_i$ and $\mathbf{H}_{j+1}^j$ are information matrices for neural network observations and IMU kinematic predictions, respectively. $\mathcal{N}$ is the set of network-based translations and uncertainties. $\mathcal{I}$ denotes the set of raw IMU measurements. The Mahalanobis norm in the objective function takes into account the correlations of the data set. 

Given the network output $\hat{\mathbf{z}}_i$, the geometric observation is only the translation $\hat{\bm{t}}_i$ between state $\mathbf{x}_i$ and $\mathbf{x}_{i+1}$. Then we can derive the matrix $\mathbf{H}_i$ from $\hat{\bm{t}}_i = \mathbf{p}_{i+1}^0 - \mathbf{p}_i^0$. The covariance $\mathbf{P}_i$ can be initialized by the network observed uncertainty $\hat{\bm{\Gamma}}_i$. Given the IMU 3D accelerations and angular velocities $\hat{\bm{a}}_t$ and $\hat{\bm{\omega}}_t$ at time $t$, the kinematic measurement $\hat{\mathbf{u}}_{j+1}^j$ and the corresponding $\mathbf{H}_{j+1}^j$ can be derived. The covariance $\mathbf{P}_{j+1}^j$ can be computed by discrete-time propagation within $\Delta t_j$~\cite{qin2018vins}. 

{\bf Loosely-coupled fusion}. At time $i$, we have obtained state $^I\hat{\bm{x}}_i$, which includes position $^I\hat{\mathbf{p}}_i^0$ from the above learned-inertial state estimation and the heading $^I\hat{\bm{q}}_i^0$ corrected by the IMU compass. $^I\hat{\bm{q}}_i^0$ is the quaternion representation of rotation. Now we would like to use these results to mitigate the temporal drift of the VI rigid transformation $^V\hat{\bm{x}}_i$ obtained by VINS~\cite{qin2018vins}, $^V\hat{\bm{x}}_i = [^V\hat{\mathbf{p}}_i^0, ^V\hat{\bm{q}}_i^0]$.

Since VINS has already estimated the rigid transformation, the state propagation is trivial in our fusion. Given a covariance matrix of sensing noise $\bm{W}$, the state covariance $\bm{P}_k$ is propagated as $\bm{P}_{k+1} = \bm{P}_{k} + \bm{W}$. Then, we derive the measurement update when taking a learned-inertial rigid transformation. We omit the details of the Kalman gain derivation for brevity. 

At this stage, we can translate the accurate rigid transformation into the scene flow of static points as follows. Give a static point $\mathbf{P}^{c_i}$, the scene flow of this point $\mathbf{f}_{c_i}^{c_j}$ from camera frame $i$ to $j$ can be written as
\begin{equation}
	\mathbf{f}_{c_i}^{c_j} = (\mathbf{I}-\hat{\mathbf{R}}_{c_i}^{c_j})\mathbf{P}^{c_i} + \hat{\mathbf{t}}_{c_i}^{c_j},
\end{equation}
where $\hat{\mathbf{R}}_{c_i}^{c_j}$ and $\hat{\mathbf{t}}_{c_i}^{c_j}$ are the rotation and translation from the rigid transformation.

\subsection{Optical-mmWave Supervision Extraction}
\label{subsec:supervision}



Our framework yields supervision signals to enable the self-supervised learning for mmWave radar's rigid transformation and scene flow. The learned model eventually provides the rigid transformation and scene flow estimation using only the radar's sparse point clouds even in smoke-filled environments. 

{\bf Rigid transformation supervision}. From \S~\ref{subsec:fusion}, we have obtained accurate odometry of the vehicle. Thus, it is trivial to compute the pseudo-ground-truth rigid transformation of the radar $\mathbf{T} \in \text{SE}(3)$ between two frames. By the Kabsch algorithm~\cite{kabsch1976solution}, we also have the rigid transformation estimation $\hat{\mathbf{T}} \in \text{SE}(3)$ between two frames. Given two radar frames $\mathcal{F}_1$, $\mathcal{F}_2$, and the rigid transformation from frame $1$ to frame $2$ $\mathbf{T}_1^2=\begin{bmatrix}
	\bm{R}_1^2 & \bm{t}_1^2 \\
	\bm{0} & 1
\end{bmatrix}.$
Then, point $\mathbf{s}_{i1} \in \mathcal{F}_1$ can be transformed into frame $2$ as
\begin{equation}
	\mathbf{s}_{i2}^\prime = \bm{R}_1^2 \mathbf{s}_i^1 + \bm{t}_1^2.
\end{equation}

Our system aims to adjust the rigid transformation estimation $\hat{\mathbf{T}}=\begin{bmatrix}
	\hat{\bm{R}} & \hat{\bm{t}} \\
	\bm{0} & 1
\end{bmatrix}$ to close to the pseudo ground-truth transformation $\mathbf{T}=\begin{bmatrix}
	\bm{R} & \bm{t} \\
	\bm{0} & 1
\end{bmatrix}$. Thus we define the loss function
\begin{equation}
	\mathcal{L}_{\text{trans}} = \frac{1}{N}\sum_{i=1}^N\left\| \left(\bm{R}\hat{\bm{R}}^\top - \mathbf{I}_3\right)\mathbf{s}_{i1} +  \bm{t} - \hat{\bm{t}} \right\|_2.
\end{equation}

{\bf Scene flow supervision}. The rigid transformation can only deduce the scene flow of static points. To fully supervise the scene flow, we need to obtain accurate scene flow of dynamic points. We first apply the dynamic point's mask from Segment Anything~\cite{kirillov2023segany} to pick out points on foreground objects, denoted as $\mathcal{D}$. Then, we adopt our previous work~\cite{fan2024enhancing}, which restores the 3D positions of dynamic visual features $\{\mathbf{C}_i\in \mathbb{R}^3\}_{i=1}^N$, $N$ is the number of points in $\mathcal{D}$. We formulate a new loss $\mathcal{L}_{\text{dyn}}$ for the scene flow supervision as follows.
\begin{equation}
	\mathcal{L}_{\text{dyn}} = \frac{1}{N}\sum_{i=1}^N\left\|\mathbf{f}_{C_i} - \hat{\mathbf{f}}_i\right\|^2,
\end{equation}
where $\mathbf{f}_{C_i}$ denotes the scene flow across two frames computed by the positions of dynamic points using~\cite{fan2024enhancing}, $\hat{\mathbf{f}}_i$ is the coarse flow from the scene flow head~\cite{ding2023hidden}.

In addition, for dynamic points in $\mathcal{D}$, we apply the loss with respect to the optical flow from images $\mathcal{L}_\text{opt}$ as 
\begin{equation}
	\mathcal{L}_\text{opt} = \frac{1}{N}\sum_{i=1}^N\left\|\mathbf{o}_{i} - \text{proj}\{\hat{\mathbf{f}}_i\}\right\|^2,
\end{equation}
where $\mathbf{o}_{i}$ the optical flow of point $i$ from~\cite{teed2020raft}, $\text{proj}\{\hat{\mathbf{f}}_i\}$ the operation that projects the radar's coarse flow to an image. 
Finally, we combine the loss of the radar point cloud itself $\mathcal{L}_\text{self}$ in~\cite{ding2022self} to formulate the overall scene flow loss, 
\begin{equation}
	\mathcal{L}_\text{flow}=\lambda_\text{opt}\mathcal{L}_\text{opt} + \mathcal{L}_\text{self} + \mathcal{L}_{\text{dyn}},
\end{equation}
where $\lambda_\text{opt}$ = 0.1 in experiments.

\label{sec:evaluation}
\section{Implementation and Evaluation}


VISC requires raw acceleration and angular velocity data from an IMU, which are not available in existing 4D radar public datasets~\cite{nuscenes, apalffy2022}. To this end, we use synthetic data generated from the Carla simulator~\cite{Dosovitskiy17} and collect real-world data by a customized platform on top of a drone. 

\begin{figure}[h]
	\centering
	\includegraphics[width=2.8in]{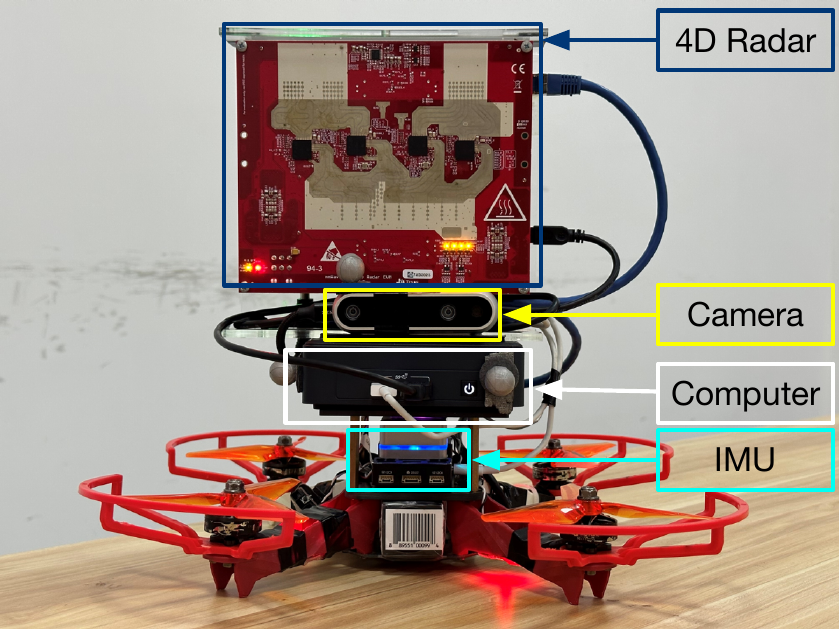}
	\caption{The key components of our customized platform include a 4D radar, an RGB camera, an IMU, and a mini-computer.}
	\label{fig:drone}
	\vspace{-3mm}
\end{figure}

{\bf Platform}. As shown in Fig.~\ref{fig:drone}, the 4D radar is cascaded by $4$ TI AWR 2243, which contains $12$ transmitting antennas and $16$ receiving antennas. The maximum sensing range is $150$ m in the MIMO model. The angle resolutions are $1.4\degree$ and $18\degree$ in azimuth and elevation directions, respectively. We use the Intel Realsense D435 camera to provide RGB images. The inertial measurements are from the integrated high-precision IMU in CUAV v5+ flight controller. All these data are sent to the Intel NUC11TNKi5 running Ubuntu 20.04 with a $2.6$ GHz Intel Core i5 CPU and a $16$ G RAM. We use the robot operating system (ROS) to establish communications among multiple sensors and temporally align their measurements. 

{\bf Dataset}. In the simulator, we mount an RGB camera, an IMU, and a 4D mmWave radar on top of a car and collect data from $8$ different scenes. This dataset includes $7119$ frames of synchronized images and point clouds, with approximately $80\%$ of the targets in each image are dynamic, and $23750$ IMU sequences. The duration of data collection is over a trajectory of $2.53$ kilometers. In real-world experiments, we collect data in outdoors (along the road in front of our campus) and indoors (in our lab). It includes $5312$ frames of synchronized images and radar point clouds for outdoor scenes, and around $1328$ frames for indoor scenes. Additionally, it consists of about $39600$ IMU sequences in outdoors, and $9900$ IMU sequences in indoors, covering distances over $2.1$ kilometers. Notably, across both indoors and outdoors, approximately $60\%$ of the objects in the images are moving.

{\bf Ground truth labelling.} Although VISC does not need the ground truth labels to train the model, we still need it to evaluate the performance. Following~\cite{ding2023hidden}, we annotate the ground truth scene flow by object detections, \ie, bounding boxes, and ground truth radar rigid transformation. In the indoor scene, we use NOKOV motion capture system to obtain the ground truth rigid transformation. In the outdoor scene, we take VINS-Fusion to fuse images, IMU and GPS measurements to obtain the ground truth rigid transformation. 

{\bfseries Metrics.} Following~\cite{ding2022self}, we assess the performance of the three sub-tasks by six metrics. We evaluate scene flow estimation by 1) the average end-point error ($EPE$ meters) that quantifies the mean distance error between the ground truth and the predicted scene flow vectors; 2) accuracy ratios, denoted by $AccS$ and $AccR$, are used to measure the ratio of points meeting strict or relaxed conditions, \ie, $EPE < 0.05$ or $0.1$ meters, respectively. The metrics for rigid transformation estimation include the relative translation error ($RTE$) and the relative angular error ($RAE$).


{\bfseries Baselines.} To our knowledge, RAFlow~\cite{ding2022self} and CMFlow~\cite{ding2023hidden} are the most related works. RAFlow leverages the measured radial velocity of radar point clouds to provide implicit supervision signals to learn the scene flow model. CMFlow needs the data from LiDAR and RTK-GPS. In our setting, we use the point cloud generated from the camera's depth map and VINS-Fusion's rigid transformation to run CMFlow.





\subsection{Performance Evaluation}
\subsubsection{Overall Results}
\begin{figure*}[h]
	\centering
	\includegraphics[width=6in]{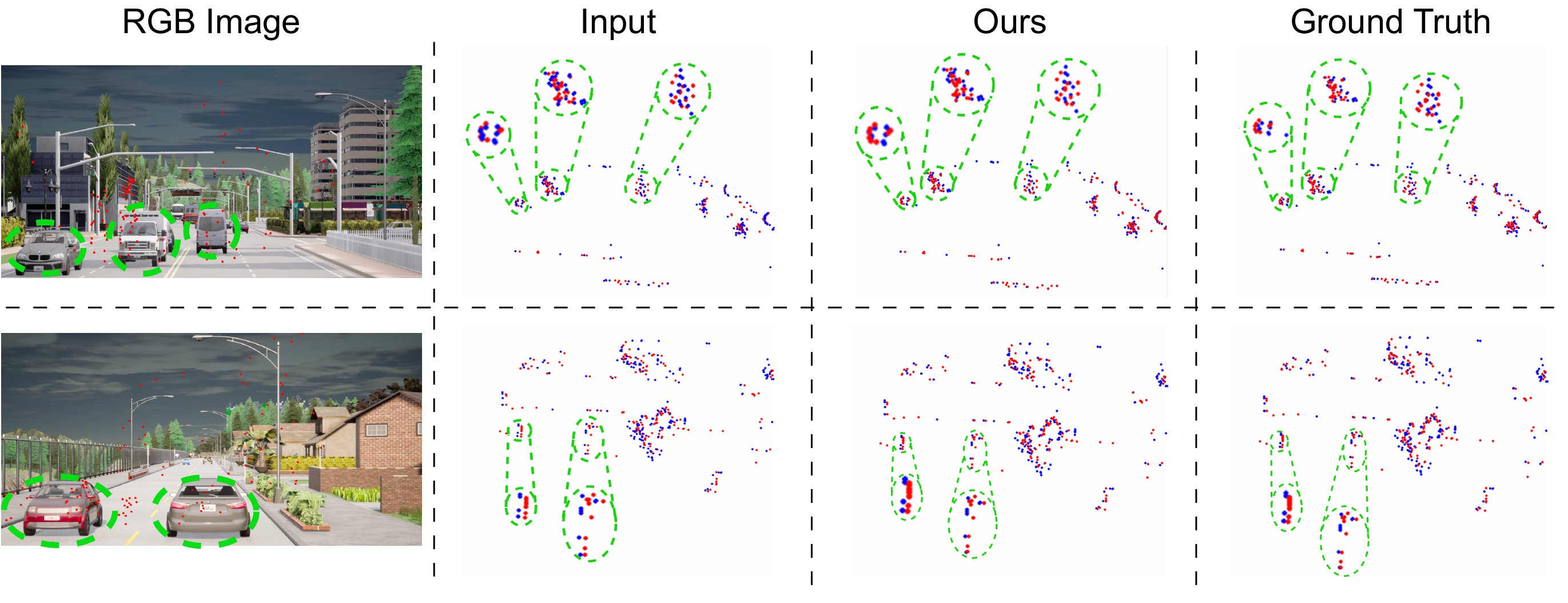}
	\caption{Qualitative scene flow results in two scenarios. From left to right: 1) RGB images with the projection of radar points; 2) two consecutive input point clouds, the first one (red) and the second one (blue); 3) the first point cloud warped by the predicted scene flow and the second point cloud; 4) the first point cloud warped by ground truth scene flow and the second point cloud. We highlight dynamic objects in green circles and zoom in on them.}
	\label{fig:scene_flow}
\end{figure*}

We compare the performance of VISC with the baselines. As shown in Table~\ref{tab:overall}, our solution shows competitive performance to CMFlow, demonstrating the VISC's effectiveness that uses low-cost visual and inertial sensors. Specifically, as expected, VISC significantly outperforms RAFlow, which is a radar-based self-supervised method, using both the synthetic data and real-world data. On the other hand, the performance of VISC is almost identical to CMFlow. When using the synthetic data, we can see a slight decrease in $AccS$ ($15.2\%$) and $AccR$ ($3.2\%$) while the mean $EPE$ is bounded. When using real-world data, VISC shows inferior performance. We believe that it is due to the lack of temporal calibration for the 4D radar. Nevertheless, VISC still shows competitive results to CMFlow. Fig.~\ref{fig:scene_flow} shows the qualitative scene flow results in synthetic scenarios.




\begin{table}[ht]
	\centering
	\caption{Method Comparison}
	\label{tab:overall}
	\begin{tabular}{p{2.38cm}|p{0.8cm}|p{0.8cm}|c|c}
		\hline
		\textbf{Method} & \textbf{Super-vision} & \textbf{EPE [m]$\downarrow$} & \textbf{AccS$\uparrow$} & \textbf{AccR$\uparrow$} \\
		\hline
		RAFlow (synth.)& Self & 0.224 & 0.286 & 0.525 \\
		RAFlow (real.)& Self & 0.283 & 0.224 & 0.475 \\
		CMFlow (synth.)& Cross & 0.136 & 0.481 & 0.774 \\
		CMFlow (real.)& Cross & 0.169 & 0.402 & 0.703 \\
		VISC (synth.)& Cross & {\bf 0.139} & {\bf 0.458} & {\bf 0.749} \\
		VISC (real.) & Cross & 0.172 & 0.375 & 0.711 \\
		\hline
	\end{tabular}
\end{table}

\subsubsection{Testing in Smoke-filled Environments}

VISC aims to provide more reliable scene flow estimation in vision-crippled weathers using mmWave radar. Therefore, we examine the effectiveness of VISC in smoke-filled environments in both simulations and real-world indoor experiments. We simulate foggy environments in Carla and using a smoke machine to produce different smoke densities in our lab. Experimental results in Table.~\ref{tab:Fog} indicate that under various levels of smoke densities in indoors, VISC demonstrates the effectiveness in that the accuracy in scene flow estimation is slightly better than CMFlow, which uses costly LiDAR data for the training. 

\begin{table}[ht]
	\centering
	\caption{Comparison in Smoke-filled Indoor Environments}
	\label{tab:Fog}
	\begin{tabular}{p{3.5cm}ccc}
		\hline
		\textbf{Smoke Density} & \textbf{EPE [m]$\downarrow$} & \textbf{AccS$\uparrow$} & \textbf{AccR$\uparrow$} \\
		\hline
		CMFlow-$30\%$ (synth.)& 0.259 & 0.264 & 0.501 \\
		CMFlow-$50\%$ (synth.)& 0.273 & 0.260 & 0.495 \\
		CMFlow-$70\%$ (synth.)& 0.322 & 0.211 & 0.439 \\
		VISC-$30\%$ (synth.)& {\bf 0.194} & {\bf 0.323} & {\bf 0.587} \\
		VISC-$50\%$ (synth.)& 0.225 & 0.281 & 0.516 \\
		VISC-$70\%$ (synth.)& 0.275 & 0.259 & 0.490 \\
		CMFlow-Light (real-w.)& 0.280 & 0.230 & 0.485 \\
		CMFlow-Med. (real-w.)& 0.298 & 0.218 & 0.451 \\
		CMFlow-Heavy (real-w.)& 0.355 & 0.135 & 0.227 \\
		VISC-Light (real-w.)& {\bf 0.216} & {\bf 0.295} & {\bf 0.568} \\
		VISC-Med. (real-w.)& 0.255 & 0.275 & 0.510 \\
		VISC-Heavy (real-w.)& 0.305 & 0.215 & 0.448 \\
		\hline
	\end{tabular}
\end{table}

\subsubsection{Performance on Rigid Transformation Estimation}
\label{subsubsec:subtasks}

\begin{figure}[h]
	\centering
	\includegraphics[width=3in]{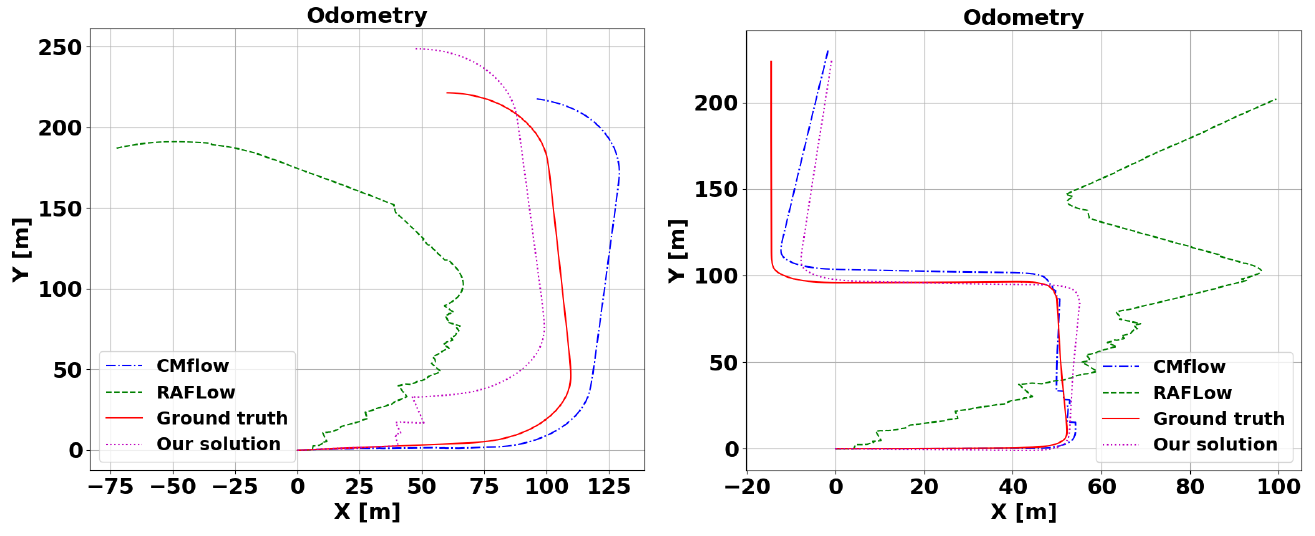}
	\caption{The trajectories are plotted based on accumulated rigid transformation for two scenarios.} 
	\label{fig:pose}
	\vspace{-3mm}
\end{figure}

In addition to scene flow estimation, we evaluated the rigid transformation estimation. We obtain rigid transformation between consecutive radar frames under the constraints of VI-SLAM and the kinematic-learned sensor fusion module. In Table.~\ref{tab:Pose}, we evaluate the rigid transformation estimation, and under the constraints of the VI-SLAM and kinematic-learned sensor fusion module, favorable rigid transformation estimation results can be obtained. Using the accumulated rigid transformation, we visualize the trajectories in Fig.~\ref{fig:pose} for two scenarios. As CMFlow outperforms VISC in terms of RTE and RAE metrics, it is expected that CMFlow will produce more accurate trajectories. However, VISC provides much better long-term trajectory estimation with the kinematic-learned sensor fusion module. VISC has much lower accumulated drifts than CMFlow, even though VISC uses low-cost VI sensors. 
\begin{table}[!htb]
      \centering
        \caption{Rigid Transformation Estimation}
        \label{tab:Pose}
       \begin{tabular}{p{0.5cm}|p{2.4cm}|p{1.2cm}|p{1.2cm}}
		\hline
		\textbf{VIO} & \textbf{Kine.-learned. Fus.} & \textbf{RTE [m]$\downarrow$} & \textbf{RAE [\degree]$\downarrow$} \\
		\hline
		&  & 0.185 & 0.203 \\
		\checkmark & & 0.116 & 0.180 \\
		\checkmark & \checkmark & {\bf 0.076} & {\bf 0.147} \\
		\hline
		\end{tabular}
\end{table}

\label{sec:conclusion}
\section{Conclusion}
This paper presents VISC, a visual-inertial (VI) self-supervised learning framework for mmWave radar scene flow estimation, moving toward crowd-sourcing VI training data. The core of our design lies in fusing kinematic model-based ego-motions with neural network learned results to obtain drift-free rigid transformation, ensuring the reliable supervision to the radar rigid transformation and the scene flow of static points. Then, our proposed optical-mmWave supervision extraction module extracts the supervision signals of rigid transformation and scene flow. With the data collected from the Carla simulator and our customized sensor platform, our extensive experiments show that VISC can even outperform the SOTA solutions using high-cost LiDARs. In future, we would like to explore the sensing resolution of mmWave radar, especially on the elevation angle, to boost the robustness and accuracy of radar perceptions.




{
    \small
    \bibliographystyle{IEEEtran}
    \bibliography{main}
}


\end{document}